\documentclass{article}

\usepackage{PRIMEarxiv}
\usepackage{float} 

\usepackage[utf8]{inputenc} 
\usepackage[T1]{fontenc}    
\usepackage{hyperref}       
\usepackage{url}            
\usepackage{booktabs}       
\usepackage{amsfonts}       
\usepackage{nicefrac}       
\usepackage{microtype}      
\usepackage{lipsum}
\usepackage{graphicx}       
\usepackage{amsmath}       
\usepackage{listings}
\usepackage{xcolor}
\usepackage{fancyhdr}       
\usepackage{graphicx}       
\graphicspath{{media/}}     
\usepackage{caption}
\usepackage{multirow}
\usepackage{soul}
\usepackage{enumitem}
\setlist[itemize]{leftmargin=*,itemsep=0pt, topsep=0pt, parsep=0pt}
\usepackage{pdfpages}

\usepackage[most]{tcolorbox} 
\usepackage{fancyvrb}        
\usepackage{xcolor}          
\usepackage{soul}            

\usepackage{makecell}
\usepackage{tabularx}

\definecolor{aigold}{RGB}{244,210, 1} 
\definecolor{aigreen}{RGB}{210,244,211} 

\sethlcolor{aigreen}

\definecolor{aired}{RGB}{255,180,181}

\definecolor{aigold}{RGB}{255,180,181}

\definecolor{aiblue}{RGB}{173,216,230} 
 
\definecolor{lightred}{rgb}{1,0.9,0.9} 

\newcounter{textbox}
\tcbset{
  custombox/.style={
    width=512.18663pt,
    top=3pt,
    fonttitle=\bfseries\small\sffamily, 
    colback=white,
    colframe=black,
    colbacktitle=black,
    boxrule=0.5pt, 
    colback=white, 
    enhanced,
    rounded corners, 
    center,
    boxed title style={
            sharp corners,
            size=small,
            colframe=blue!75!black, 
        },
    attach boxed title to top left={yshift=-0.1in,xshift=0.15in},
    boxed title style={boxrule=0pt,colframe=white,},
    before=\setlength{\baselineskip}{-2pt}, 
    before upper=\setlength{\baselineskip}{8pt}, 
    after=\setlength{\baselineskip}{8pt}, 
  }
}
\newtcolorbox{LLMbox}[2][]{custombox, width=\textwidth, title=#2, #1}

\newtcbox{\mybox}[1][green]{on line,
arc=0pt,outer arc=0pt,colback=#1!10!white,colframe=#1!50!black,
boxsep=0pt,left=0pt,right=0pt,top=0pt,bottom=0pt,
boxrule=0pt,bottomrule=0pt,toprule=0pt}

\tcbset{
  customboxmultipage/.style={
    width=\textwidth, 
    top=2pt, 
    bottom=2pt, 
    left=2pt, 
    right=2pt, 
    fonttitle=\bfseries\small\sffamily, 
    colback=white, 
    colframe=black, 
    boxrule=0.5pt, 
    breakable, 
    rounded corners, 
    center, 
    boxed title style={
      sharp corners,
      size=small,
      colframe=blue!75!black, 
      colback=white, 
    },
    attach boxed title to top left={
      yshift=-0.1in, 
      xshift=0.15in  
    },
    boxed title style={
      boxrule=0pt, 
      colframe=white, 
    },
    enhanced, 
    nobeforeafter, 
  }
}

\newtcolorbox{LLMboxmultipage}[2][]{customboxmultipage,title=#2,#1}

\lstdefinelanguage{json}{
    basicstyle=\ttfamily\scriptsize, 
    numbers=left,
    numberstyle=\tiny\color{gray}, 
    stepnumber=1,
    numbersep=10pt, 
    showstringspaces=false,
    breaklines=true,
    frame=none,
    backgroundcolor=\color{white},
    literate=
     *{0}{{{\color{blue}0}}}{1}
      {1}{{{\color{blue}1}}}{1}
      {2}{{{\color{blue}2}}}{1}
      {3}{{{\color{blue}3}}}{1}
      {4}{{{\color{blue}4}}}{1}
      {5}{{{\color{blue}5}}}{1}
      {6}{{{\color{blue}6}}}{1}
      {7}{{{\color{blue}7}}}{1}
      {8}{{{\color{blue}8}}}{1}
      {9}{{{\color{blue}9}}}{1}
      {:}{{{\color{red}:}}}{1}
      {,}{{{\color{red},}}}{1}
      {\{}{{{\color{brown}\{}}}{1}
      {\}}{{{\color{brown}\}}}}{1}
      {[}{{{\color{brown}[}}}{1}
      {]}{{{\color{brown}]}}}{1},
}

\lstdefinelanguage{yaml}{
  basicstyle=\ttfamily\footnotesize, 
  morekeywords={slices, sources, model, layer_range, merge_method, base_model, parameters, t, filter, value, dtype}, 
  keywordstyle=\color{blue},
  comment=[l]{\#},
  commentstyle=\color{gray},
  morestring=[b]',
  morestring=[b]",
  stringstyle=\color{orange},
  sensitive=true,
}

\title{
\bf{Generative Artificial Intelligence Extracts Structure-Function Relationships from Plants for New Materials}
\thanks{\textit{\underline{Citation}}: 
\textbf{R.K. Luu, et al., Title. Pages.... DOI:000000/11111.}} 
}

\author{
    Rachel K. Luu \\
  Department of Materials Science and Engineering \\
   Laboratory for Atomistic and Molecular Mechanics (LAMM) \\
  Massachusetts Institute of Technology\\
  Cambridge, MA, USA\\
  https://orcid.org/0000-0002-7821-934X
   \And
    Jingyu Deng \\
  School of Materials Science and Engineering\\
  Centre for Cross Economy Global\\
  Nanyang Technological University\\
  Singapore\\
  https://orcid.org/0000-0003-2765-1906
   \And
     Mohammed Shahrudin Ibrahim \\
  School of Materials Science and Engineering\\
  Centre for Cross Economy Global\\
  Nanyang Technological University\\
  Singapore\\
  https://orcid.org/0000-0002-3183-8581
   \And
     Nam-Joon Cho \\
  School of Materials Science and Engineering\\
  Centre for Cross Economy Global\\
  Nanyang Technological University\\
  Singapore\\
  https://orcid.org/0000-0002-8692-8955
   \And
   Ming Dao \\
    Department of Materials Science and Engineering \\
    Massachusetts Institute of Technology \\
  Cambridge, MA, USA\\
  https://orcid.org/0000-0001-5372-385X
   \And
   Subra Suresh \\
    Department of Materials Science and Engineering \\
    Massachusetts Institute of Technology \\
  Cambridge, MA, USA\\
  http://orcid.org/0000-0002-6223-6831
   \And
  Markus J. Buehler \\
  Department of Civil and Environmental Engineering\\
  Department of Mechanical Engineering \\  
  Center for Computational Science and Engineering \\
  Schwarzman College of Computing \\  
  Laboratory for Atomistic and Molecular Mechanics (LAMM) \\
  Massachusetts Institute of Technology \\
  Cambridge, MA, USA\\ 
  https://orcid.org/0000-0002-4173-9659\\
  \\
  \texttt{mbuehler@MIT.EDU} 
}

\begin{document}
\maketitle

\begin{abstract}
Large language models (LLMs) have reshaped the research landscape by enabling new approaches to knowledge retrieval and creative ideation. Yet their application in discipline-specific experimental science, particularly in highly multi-disciplinary domains like materials science, remains limited. We present a first-of-its-kind framework that integrates generative AI with literature from  hitherto-unconnected fields such as plant science, biomimetics, and materials engineering to extract insights and design experiments for materials. We focus on humidity-responsive systems such as pollen-based materials and \textit{Rhapis excelsa} (broadleaf lady palm) leaves, which exhibit self-actuation and adaptive performance. Using a suite of AI tools, including a fine-tuned model (BioinspiredLLM), Retrieval-Augmented Generation (RAG), agentic systems, and a Hierarchical Sampling strategy, we extract structure-property relationships and translate them into new classes of bioinspired materials. Structured inference protocols generate and evaluate hundreds of hypotheses from a single query, surfacing novel and experimentally tractable ideas. We validate our approach through real-world implementation: LLM-generated procedures, materials designs, and mechanical predictions were tested in the laboratory, culminating in the fabrication of a novel pollen-based adhesive with tunable morphology and measured shear strength, establishing a foundation for future plant-derived adhesive design. This work demonstrates how AI-assisted ideation can drive real-world materials design and enable effective human–AI collaboration.
\end{abstract}


\section{Introduction}
 
Generative AI tools such as large language models (LLMs) provide powerful pathways to accelerate research and development, especially creating novel materials with never-before-seen design principles. Previous studies have explored fine-tuning foundational LLMs with domain-specific knowledge\cite{Lu2024Fine-tuningCapabilities_updated} such as in medicine~\cite{Chen2024CanDomain}, chemistry \cite{VanHerck2025AssessmentApplications}, coding \cite{Roziere2023CodeCode}, and, most relevant to this study, biological materials science with the BioinspiredLLM model\cite{Luu2023BioinspiredLLM:Materials}. However, while these domain-specific LLMs have shown promise in knowledge retrieval, their direct application to problem solving and creative ideation in specific experimental research remains limited. Building on these foundations, we advance the role of generative AI in science by demonstrating its integration not only in hypothesis generation but also in experimental design and validation as well as material fabrication, a critical step toward bridging AI-driven ideation with scientific practice and real-world experimental challenges.

This work deals with a subset of the field of biological materials by focusing on plant science, and the connections among structure, properties, function, and performance, invoking the analogy with modern materials science. The shape-morphing properties of plants have been extensively studied for generations \cite{Speck2011PlantMechanics, Gibson2012TheMaterials}, exhibiting multifunctional surfaces \cite{Barthlott2017PlantInnovations}, intricate morphologies \cite{Zhao2018OnPlants}, and inspiring shape-morphing abilities \cite{Bar-On2014StructuralTissues}. In particular, research in plant mechanics has inspired numerous bioinspired applications \cite{Burgert2009ActuationDevices, Sadeghi2016ACapabilities}. The mechanics of plants, governed largely by environmental stimuli, has been widely explored in response to changes in humidity, temperature, light, and touch \cite{Braam2005InStimuli, Schlanger2024ThePlants}.

\begin{figure}[h]
    \centering
    \includegraphics[width=1\linewidth]{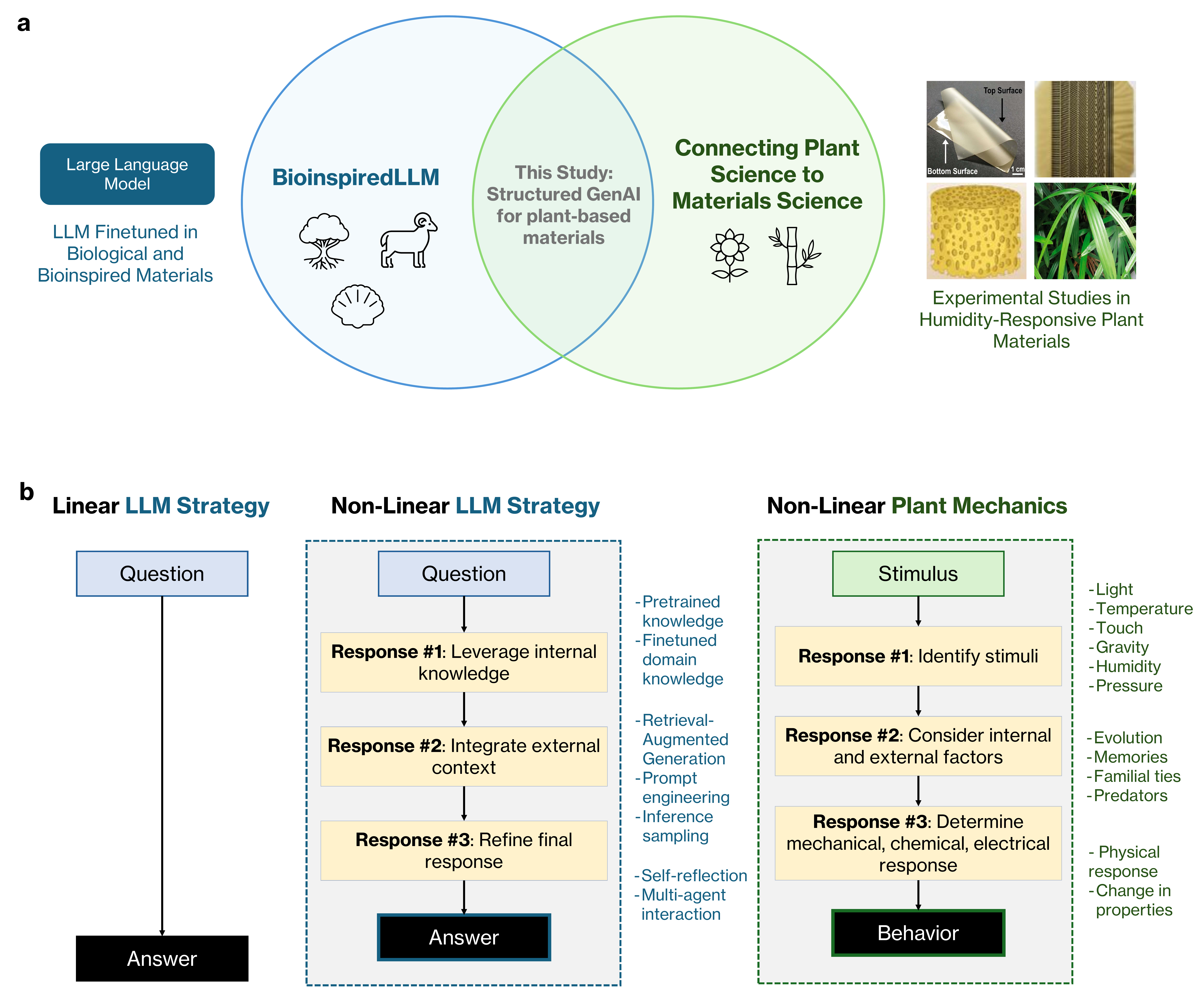}
    \caption{Study overview \& Linearity vs. Non-Linearity of LLMs and Plants a) Scope of this work, regarding the intersection of Generative AI studies (BioinspiredLLM) with Experimental Mechanics studies (plant mechanics, particularly humidity-responsive materials) b) Comparison of Linear vs. Non-Linear strategies in LLM usage as well as Non-Linear phenomena found in plant materials.}
    \label{fig:Fig_1}
\end{figure}

While covering broad interdisciplinary topics, we confine this study specifically to recent studies of hygro-morphing plant-based materials, in order to get into some significant depth. Due to their inherent humidity responsiveness, we focus attention on recent advances in creating non-allergenic pollen-based microgels \cite{Zhao2020ActuationPollen, Zhao2021DigitalMaterials} and pollen cryogels \cite{Deng2024Plant-BasedControl}, as well as on the mechanics of \textit{Rhapis excelsa} (broadleaf lady palm) leaves \cite{Guo2024Dehydration-inducedLeaves}, all of which autonomously alter their morphology and properties in response to moisture changes. These systems were selected not only for their responsive behavior and relevance to bioinspired design, but also because our proximity to their experimental development provides access to detailed domain knowledge and constraints. Notably, they enable functions such as self-actuation and adaptive performance that are difficult to achieve with synthetic materials, making them compelling platforms for AI-assisted materials exploration. While other classic plant systems—such as pine cones \cite{Zhang2022UnperceivableCones}—are also known for humidity responsiveness, we prioritized systems where we could engage directly with the experiments and underlying data to enable deeper integration with AI-driven analysis. Brief summaries of each paper are included in Table \ref{tab:papers_summary}. These works, which will henceforth be generally referred to as the plant studies or plant literature, offer a realistic yet focused framework for exploring how generative AI tools can be integrated into highly specific materials science and engineering domains, Fig \ref{fig:Fig_1}a.

\begin{table}[h]
\centering
\small
\setlength{\tabcolsep}{4pt}
\begin{tabular}{p{4.2cm} p{2.2cm} p{2.5cm} p{7cm}}
\toprule
\textbf{Article Title} & \textbf{First Author, Year} & \textbf{Plant Material} & \textbf{Brief Summary} \\
\midrule
Actuation and locomotion driven by moisture in paper made with natural pollen\cite{Zhao2020ActuationPollen} 
& Zhao, 2020 
& Pollen paper 
& Introduces a sustainable actuator sheet material made from natural pollen capable of humidity-driven self-actuation and shape reconfiguration. \\

Digital printing of shape-morphing natural materials\cite{Zhao2021DigitalMaterials} 
& Zhao, 2021 
& Pollen paper 
& Follow-up study implementing programmable shape evolution in pollen paper through digital printing and surface coatings. \\

Plant-Based Shape Memory Cryogel for Hemorrhage Control\cite{Deng2024Plant-BasedControl} 
& Deng, 2024 
& Pollen cryogels 
& Introduces a biosafe pollen cryogel with rapid humidity-triggered shape-memory properties and effective hemostatic performance, providing a sustainable and low-energy solution for treating deep noncompressible wounds without harmful crosslinkers. \\

Dehydration-induced corrugated folding in \textit{Rhapis excelsa} plant leaves\cite{Guo2024Dehydration-inducedLeaves} 
& Guo, 2024 
& Broadleaf lady palm (\textit{Rhapis excelsa}) leaves 
& Investigates the dehydration-induced corrugated folding mechanisms in \textit{R. excelsa} leaves and applies these insights to biomimetic soft machines and humidity-responsive devices. \\
\bottomrule
\end{tabular}
\caption{Summary of plant experimental mechanics studies relevant to this work.}
\label{tab:papers_summary}
\end{table}

These plant studies provide not only rich insights from Nature, but also offer a compelling model for how plant-derived materials respond to environmental complexity. Inspired by this dynamic behavior, one of our overarching aims is to develop more non-linear implementations of LLMs. Moving beyond the conventional approach of treating LLM inference as a single-shot question-answer mechanism \cite{Luu2024LearningDesign}, instead, LLM inference should be designed to robustly integrate multiple sources of information, akin to how plant mechanics respond dynamically to external stimuli \cite{Schlanger2024ThePlants}, as depicted in Fig. \ref{fig:Fig_1}b. In addition, LLMs are uniquely capable of processing vast and diverse bodies of knowledge across disciplines at a scale far beyond human capacity. This ability allows them to identify novel connections, such as those between plant science and materials science, that would otherwise remain unrecognized. 

Yet despite their promise, some early generation LLMs were limited by unreliable or flawed outputs\cite{HuangLei2025AQuestions,Perkovic2024HallucinationsChallenges}. These shortcomings often arise from the simple single-shot inference methods used in off-the-shelf LLMs. To address this, we leverage a broader generative AI toolkit that includes domain-specific fine-tuning, \cite{Lu2024Fine-tuningCapabilities_updated, Luu2023BioinspiredLLM:Materials}, reasoning models\cite{Buehler2024PRefLexOR,Buehler2025GPReFLexOR,Buehler2025GA}, Retrieval-Augmented Generation (RAG) \cite{Lewis2020Retrieval-AugmentedTasks}, and agentic systems \cite{Buehler2024GenerativeDesign, Ni2023MechAgents:Knowledge, Ghafarollahi2024RapidSystems, Ghafarollahi2024ProtAgents:Learning, Ghafarollahi2024SciAgents:Reasoning, Ghafarollahi2024AtomAgents:Intelligence} to develop a more robust and structured approach to AI-driven materials discovery. In addition, we introduce a new inference technique termed Hierarchical Sampling, which involves generating a wide pool of candidate outputs through high-temperature sampling and then iteratively refining or filtering them based on structured criteria. This multi-stage process enhances the depth and diversity of LLM outputs, enabling more reliable and insightful ideas for scientific discovery. While some recent approaches introduce non-linear elements, such as iterative prompting, chain-of-thought reasoning, or tool-augmented agents, these methods often lack structured mechanisms for idea selection or refinement. Hierarchical Sampling addresses this gap by explicitly organizing the inference process into divergent and convergent stages, improving both the breadth and quality of generated outputs.

We hypothesize that by integrating a structured AI framework, including fine-tuned models, RAG, agentic systems and Hierarchical Sampling, LLMs can rapidly and at scale produce more diverse, insightful, and scientifically actionable outputs. We test this hypothesis through both qualitative and quantitative analyses of LLM-generated outputs, and crucially, through experimental validation. Broadly, the goal of this work is to explore how and to what extent structured generative AI systems, in collaboration with human researchers, can support and accelerate scientific research, particularly during early-stage ideation and experimental design. By grounding these tools in the domain of plant-based materials, we simultaneously demonstrate how such systems can help uncover novel design directions and illuminate structure–property relationships that may inform future bio-derived and bioinspired applications. This connection between AI-guided ideation and physical realization highlights the transformative potential of generative AI in materials research.

\section{Results and discussion}
\subsection{Design of the generative AI system}
\label{sec:DesignofGenAi}

The generative AI system developed in this study integrates a fine-tuned language model, BioinspiredLLM, with a RAG pipeline linked to the plant literature. To enable robust scientific ideation, we further incorporate agentic systems and introduce a new structured sampling method called Hierarchical Sampling, guided by two newly developed protocols designed to enhance the creativity, relevance, and executability of AI-generated outputs.

While LLMs have shown promise in imaginative tasks such as creative writing \cite{Gomez-Rodriguez2023AWriting} and role-playing scenarios \cite{Zhao2024AssessingModels}, their potential in scientific research - where creativity must be balanced with rigorous technical reasoning - has not been fully realized. Much of the discourse around LLM creativity draws from Boden’s cognitive framework \cite{Boden2004TheMechanisms}, which defines creativity through factors like novelty, surprise, and value. While these criteria can provide a theoretical foundation, they do not account entirely for the structured reasoning required in scientific inquiry. 

A key challenge in applying LLMs to scientific idea generation is that their outputs are fundamentally constrained by probabilistic token selection, where a token refers to a basic unit of text (such as a word fragment or character string) used during language model generation. Even when increasing randomness—via the temperature parameter, which raises the entropy of the model’s output distribution—certain tokens still remain statistically favored, limiting the diversity of generated ideas. To address this, we introduce Hierarchical Sampling,  a structured inference method that guides the model through staged reasoning and refinement. In practice, this involves generating a large pool of outputs in an initial divergent phase, followed by one or more filtering or synthesis stages that iteratively refine the outputs into more targeted, high-quality ideas. The name draws inspiration from the hierarchical organization found in natural materials, where structure and function emerge across multiple scales, and reflects our system's layered approach to idea generation and evaluation. Inspired by Guilford’s Structure of Intellect theory \cite{Guilford1956TheIntellect}, this method follows a divergent–convergent thinking framework, complementing conventional convergent thinking (identifying a single best solution) with divergent thinking (generating a broad set of possibilities).

Therefore, as shown in Fig.\ref{fig:Fig_2}a, our system leverages multiple interconnected tools and strategies, compared with traditional linear, single-shot outputs. Two major protocols were developed: 1. Idea Mining, which generates and refines hypotheses or material design concepts, and 2. Procedure Design, which translates high-level ideas into implementable experimental steps for laboratory fabrication. 

\subsubsection{Idea Mining}
\begin{figure}[h]
    \centering
    \includegraphics[width=.95\linewidth]{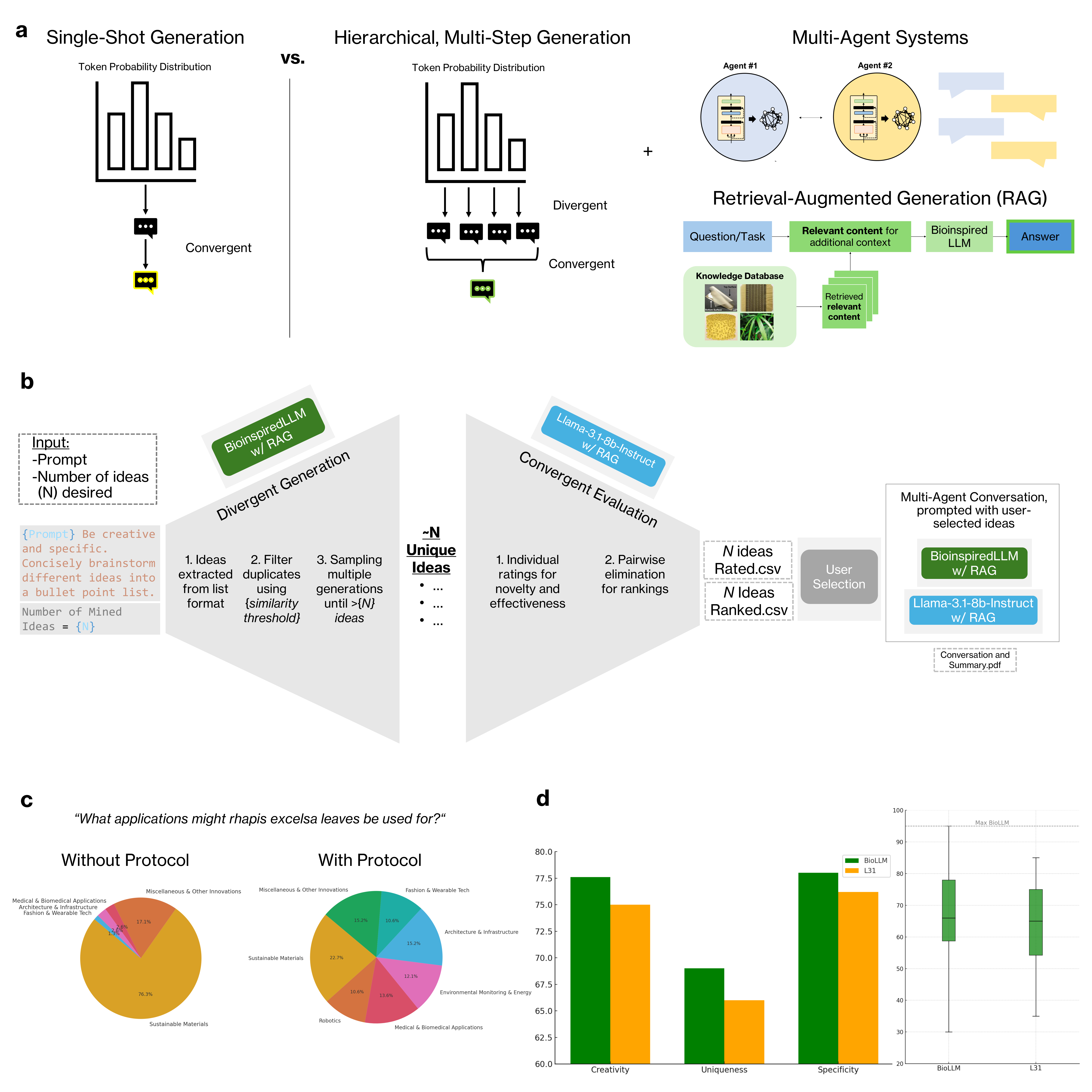}
    \caption{Our GenAI Approach \& Idea Mining Protocol a) This GenAI system shifts away from single-shot generation to hierarchical, multi-step generation sampling practices as well as employing a fine-tuned model, agentic systems and retrieval-augmented generation (RAG). b) Idea Mining Protocol, starting with an user input prompt and number of ideas desired. The two phases of Idea Mining consists of a divergent generation phase and a convergent evaluation phase which outputs human-readable files of the list of ideas, evaluated and ranked. The user then selects ideas of interest to further elaborate on them by prompting multi-agent conversation. c) Comparison of mass generation of ideas without and with the Idea Mining Protocol. d) Evaluation of 1000+ divergently generated ideas from BioinspiredLLM (BioLLM) and Llama-3.1-8b-instruct (L31) to compare effects of using a fine-tuned model versus standard foundational LLM, across a rubric regarding creativity, uniqueness, and specificity.}
    \label{fig:Fig_2}
\end{figure}

For idea generation, we implement Hierarchical Sampling as the core approach. Coined the Idea Mining protocol, as shown in Fig.~\ref{fig:Fig_2}b, the user provides a prompt, which initiates divergent sampling using BioinspiredLLM to generate a wide range of ideas. Once a target number of unique ideas (\textit{N}) is reached, the protocol transitions into a convergent phase. Here, a second model, Llama-3.1-8b-instruct \cite{TouvronLlamaModels}, evaluates each idea based on novelty and effectiveness. The output consists of two ranked lists of the \textit{N} unique ideas. Given the reasoned lists of ideas, the user may select the most promising idea(s) based on predefined criteria such as novelty, feasibility, or relevance to the materials domain. These selected ideas are then passed into a multi-agent refinement phase, where BioinspiredLLM and Llama-3.1-8b-instruct collaboratively reflect and elaborate on the concepts, outputting a summary of the conversation. 

Comparison of LLM outputs with and without the Idea Mining protocol reveals a significant reduction in diversity when the protocol is not used. As shown in Fig.~\ref{fig:Fig_2}c, single-shot generations tend to cluster around a narrow set of applications, whereas the Idea Mining protocol produces a broader distribution across multiple categories.

We also evaluated the impact of using a fine-tuned versus non-fine-tuned model for the divergent generation phase. While BioinspiredLLM was ultimately selected, we first compared its outputs to those of the base Llama-3.1-8b-instruct model. Human expert evaluation of both lists of ideas revealed that BioinspiredLLM consistently generated more specific, detailed, and unique ideas. In addition to human assessment, we conducted an automated evaluation using a third-party LLM equipped with a scoring rubric and integrated Semantic Scholar search to check for novelty. Each idea was assigned a novelty score based on the rarity of similar concepts in the literature, as determined by keyword matching and relevance-ranked results. As shown in Fig.~\ref{fig:Fig_2}d, BioinspiredLLM outperformed across all metrics. Notably, in novelty scoring, its ideas exhibited a wider range and a higher upper quartile than those of the base model, indicating both breadth and standout originality. Importantly, BioinspiredLLM achieved the highest observed novelty score, underscoring its potential to generate truly innovative concepts. Even a single breakthrough idea validates the purpose of the method: enabling discovery through diverse and high-impact ideation.

\subsubsection{Procedure Design}
To enable the generation of content with real-world applicability (such as synthesizing a new material), we developed a protocol for designing laboratory procedures, a task which requires both technical precision and creative reasoning. When using the protocol, users input a prompt and the system returns a detailed procedure.

\begin{figure}[h]
    \centering
    \includegraphics[width=1\linewidth]{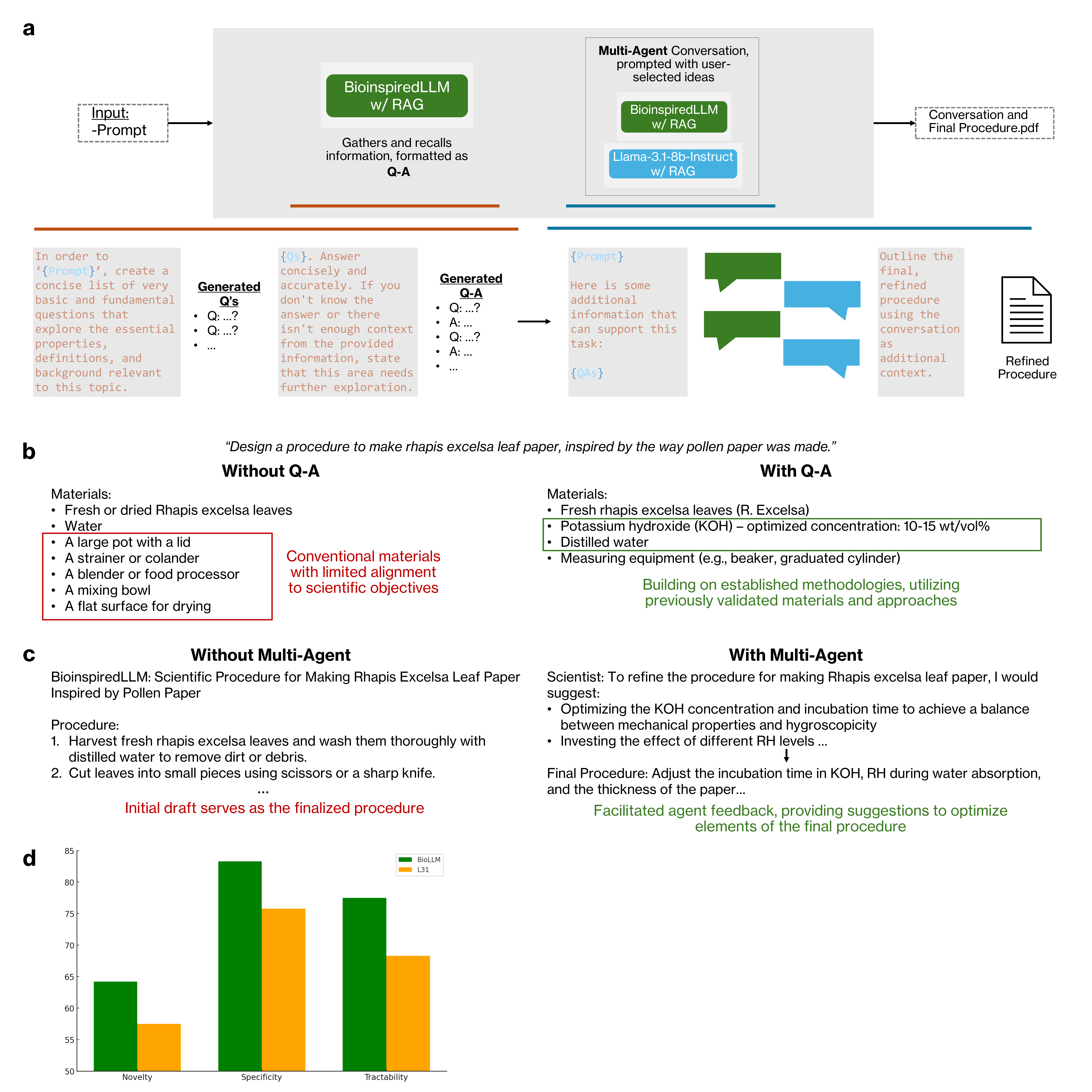}
    \caption{Procedure Design Protocol a) Procedure Design Protocol, starting with an user input prompt. The two phases of Procedure Design consists of a technical evaluation phase where one model gathers and recalls necessary information to complete task, then information aids a multi-agent conversation phase where models synthesize the design of laboratory procedure from scratch. The final, refined procedure is generated by a LLM after reviewing the conversation b) Results without and with Procedure Design protocol, highlighting differences between not including the Q-A step and including the Q-A step. c) Results without and with Procedure Design protocol, highlighting differences between not including the Multi-Agent step and including the Multi-Agent step. d) Evaluation of procedures generated using the protocol with BioinspiredLLM (BioLLM) and default Llama-3.1-8b-instruct (L31).}
    \label{fig:Fig_3}
\end{figure}

While the process appears straightforward to the user, it involves multiple steps to refine the final procedure. First, the system emphasizes technical grounding by generating relevant question–answer (Q–A) pairs based on the prompt. Then, in a second stage, a collaborative multi-agent setup takes over to synthesize the final procedure, using the Q–A content as structured context. This two-phase approach ensures that the resulting procedures are both technically sound and creatively constructed, as shown in Fig.~\ref{fig:Fig_3}a.

Due to the multi-step nature of this protocol, we observe more refined and scientifically grounded procedure outputs. As shown in Fig.~\ref{fig:Fig_3}b, omitting the Q–A generation step often leads to shallow procedures, such as recommending generic or conventional materials. In contrast, incorporating the Q–A step provides essential context which leads to more informed suggestions, consistent with past experimental studies. Additionally, the multi-agent conversation step further enhances procedural quality. Without this step (Fig.\ref{fig:Fig_3}c), a single agent generates a draft that becomes the final output with minimal iteration. When multi-agent collaboration is enabled, a second LLM contributes suggestions that are integrated into the procedure, resulting in more comprehensive outcomes.

The effectiveness of integrating the Procedure Design protocol with the system was evaluated through both blinded human review and automated scoring. In a blind assessment, human experts consistently rated procedures generated by BioinspiredLLM with the Procedure Design protocol as more promising and practically usable than those generated by an off-the-shelf Llama model. While human evaluation introduces some subjectivity, the use of standardized criteria and consistency across reviewers lends credibility to the results. To complement this qualitative evaluation, an additional automated assessment was conducted using an LLM equipped with a scoring rubric measuring novelty, specificity, and tractability. As shown in Fig.~\ref{fig:Fig_3}d, BioinspiredLLM, when guided by the Procedure Design protocol, outperformed the baseline across all evaluated metrics.

\subsection{Extracting and validating mechanistic insights}
\begin{figure}[h]
    \centering
    \includegraphics[width=1\linewidth]{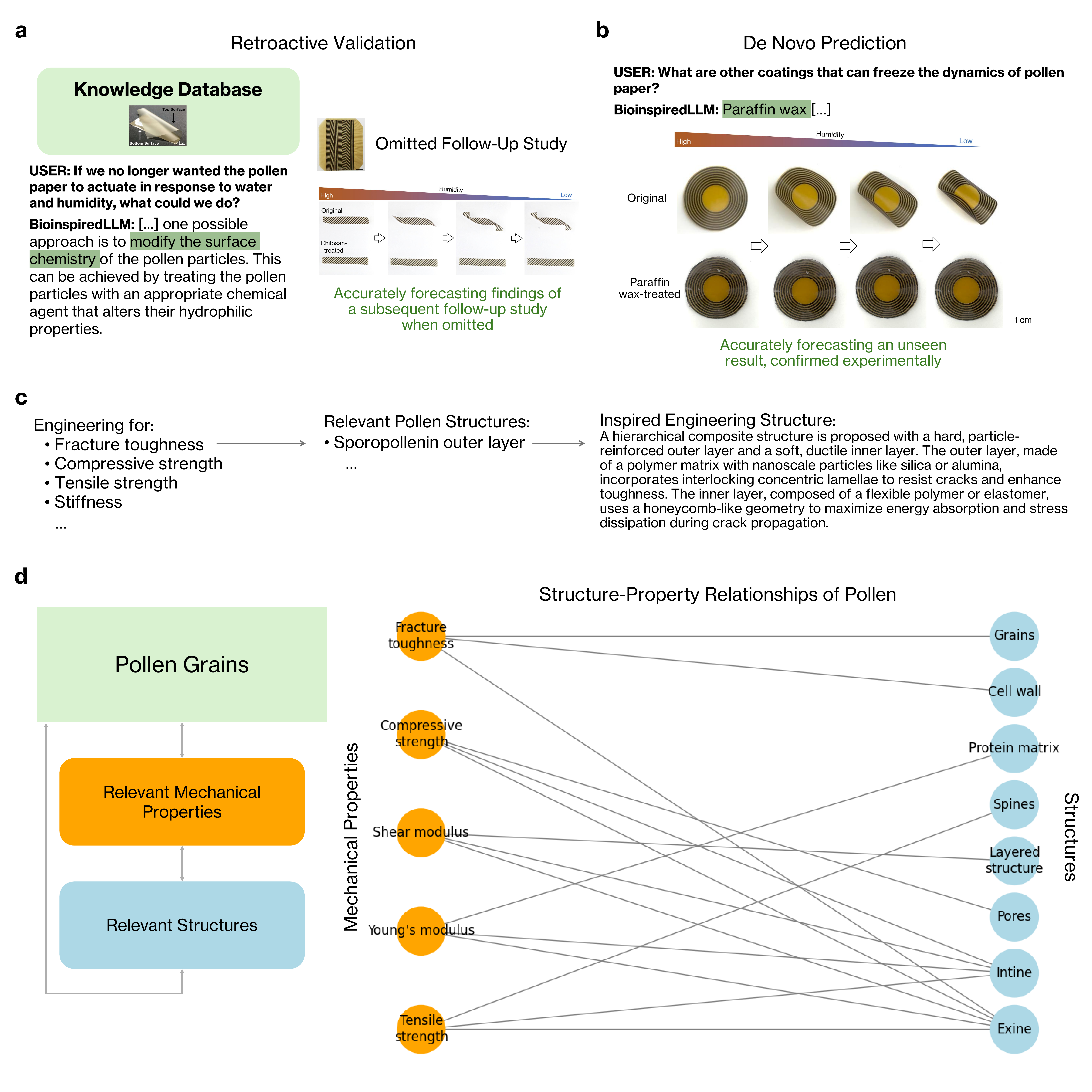}
    \caption{Extraction \& Validation of Mechanistic Insights a) System anticipating mechanical behavior via retroactive validation (predicting already proven behaviors through purposeful omission of studies) example. (Image adapted from \cite{Zhao2021DigitalMaterials}. Licensed under CC BY-NC-ND 4.0) b) System anticipating mechanical behavior via a \textit{de novo} prediction, validated experimentally in the laboratory. c) LLM-generated workflow showing the identification of mechanical properties of interest, to the correlation to relevant plant structures, and then translation of the plant structure to a bioinspired engineering design. d) Identification of Structure-Property relationships regarding pollen, represented in graph format.}
    \label{fig:Fig_4}
\end{figure}

We begin by demonstrating our system’s ability to anticipate and hypothesize mechanical behaviors and insights. As shown in Fig. \ref{fig:Fig_4}a, we supply the LLM's RAG database with only the initial work on pollen paper (Zhao 2020), purposefully omitting its follow-up study (Zhao 2021). The user then prompts BioinspiredLLM to hypothesize how to stop the actuation behavior of pollen paper, in which case the system predicts that the user can modify the surface chemistry to make the pollen paper no longer responsive to water and humidity. This response does accurately forecast the findings in the follow-up study which does use chitosan-treatment or a coating of petroleum jelly to 'freeze' the mechanical response of the pollen paper. In a \textit{de novo} prediction case, one not explicitly addressed in the existing plant literature, the user prompts the system to predict what other coatings could 'freeze' the dynamics of pollen paper, as shown in Fig.\ref{fig:Fig_4}b. The system provides a prediction of using paraffin wax. This prediction was validated experimentally in the laboratory, and it was found that paraffin wax-treated pollen paper, as predicted, did not exhibit significant observable response to changes in humidity. In both scenarios, we demonstrate the system's ability to extrapolate beyond its provided knowledge, anticipating behaviors that could inform and inspire subsequent experimental investigations. While neither prediction is highly unconventional, both reflect chemically plausible reasoning and align with known experimental strategies. These responses suggest that the model may help streamline early-stage hypothesis generation by surfacing reasonable ideas more quickly than traditional trial-and-error approaches.

Another task of particular interest to material scientists is understanding structure-property relationships, which are crucial for developing bioinspired materials. In this task, the user specifies mechanical properties of interest, and the system identifies corresponding plant structures relevant to those properties. In this example, shown in Fig. \ref{fig:Fig_4}c, the task focuses on engineering for fracture toughness. The system, using Hierarchical Sampling, is instructed to identify many relevant plant structures and then narrow down to a select few, inducing reasoning layers. For this instance, the plant structure  sporopollenin, the outer layer of pollen grains, was identified and selected after the sampling process. The top pollen structures are then passed back to the system, which translates the biological design into an description of an engineered architecture. In this case, the proposed architecture emulates sporopollenin by featuring a hard, particle-reinforced outer layer encasing a softer, ductile inner layer. Notably, the described engineering structure draws from prior research that translates biological design principles - such as interlocking lamellae - into bioinspired engineered systems \cite{Naleway2015StructuralBioinspiration}. Additionally, the description is informed by technical insights regarding mechanical relationships, linking properties like fracture toughness to crack propagation and energy absorption. While a human expert may eventually make similar connections, the system offers a distinct advantage in accelerating early-stage exploration by rapidly mapping structure-property relationships and simultaneously proposing engineering design strategies or material concepts inspired by a broader range of prior examples than a single individual might recall.

Lastly, we demonstrate how the system can be used to elucidate and organize structure–property relationships in pollen-based materials. While raw pollen grains themselves do not exhibit humidity-responsive behavior, regenerated pollen materials - such as pollen paper - do, due to the intercalation of water molecules between their layered structures. In this context, the system is prompted to identify and connect relevant mechanical properties (e.g., fracture toughness, tensile strength) with the micro- and nano-scale structural elements derived from pollen. As shown in Fig. \ref{fig:Fig_4}d, the generated graph displays nodes representing mechanical properties and pollen structures, with edges indicating the relationships between them. Notably, the system identifies subtle connections across the universal structures of pollen grains, with a predominant focus on the exine and intine. Of particular interest are the unique connections, such as the one between tensile strength and pollen spines, which mirrors relationships observed in other biological materials like sponge spicules and porcupine quills - both linked to tensile strength due to their similar aspect ratios \cite{Chen2012BiologicalDesigns, Meyers2013StructuralConnections}. Additionally, the absence of certain connections provides valuable insights. For example, the connection between fracture toughness and the exine, but not the intine, suggests that the exine (the outer surface) plays a more critical role in protecting against surface cracking. These structure-property maps offer a structured lens for understanding pollen-derived materials and guiding future design directions. While this graph was generated for the structural elements of native pollen grains, similar structure-property maps can be developed for processed systems such as pollen paper with minor changes.

\subsection{Generating ideas and  fabricating materials designs}
To generate a range of novel ideas for plant materials research prompts, we employed the Idea Mining protocol. Table \ref{tab:idea_mining_results} presents the top ideas extracted for a range of queries related to pollen-based materials and \textit{Rhapis excelsa} leaf mechanics by using the protocol. The resulting ideas demonstrate the system’s ability to generate novel yet technically grounded concepts - many of which are suitable for experimental validation and reflect the complexity and specificity of plant materials research. Table \ref{tab:idea_mining_agents} shows an example of one of these ideas, applied in a multi-agent context, and depicts the resulting summary which provides further elaboration on the specific idea. 

\begin{table}[h]
\centering
\small 
\setlength{\tabcolsep}{5pt} 
\begin{tabular}{p{6cm} p{10cm}}
\toprule
\textbf{Prompt} & \textbf{Ideas} \\
\midrule
Besides cryogels and paper, what are other specific forms that pollen grains can be used in? &
\begin{itemize}
  \item Fillers in bioplastics
  \item Photonics for light harvesting
  \item Textiles and fibers
\end{itemize} \\

What applications might \textit{Rhapis excelsa} leaves be used for? &
\begin{itemize}
  \item Water filtration systems
  \item Environmental monitoring and sensors
  \item Adaptive, smart building materials
  \item Irrigation systems for agriculture
\end{itemize} \\

Beyond hemorrhage control, where else could the high absorption properties of pollen-based cryogels be effectively applied? &
\begin{itemize}
  \item Biocompatible drug delivery
  \item Biosensors
  \item Environmental remediation (oil spills, heavy metals)
  \item Skin graft scaffolds
  \item Capture and removal of microplastics
\end{itemize} \\

Instead of the striped pattern printed on the pollen paper, what other patterns could induce interesting dynamic folding? &
\begin{itemize}
  \item Nature-inspired patterns (leaf venation)
  \item Fractals
  \item Combination or gradient of multiple patterns
  \item Spiral patterns
  \item Origami folding
\end{itemize} \\

What are shape-morphing engineering applications that can implement printed pollen paper? For each idea describe briefly what the printed pattern on the pollen paper should look like. &
\begin{itemize}
  \item Smart textiles for athletic wear (patterned microscale channels for moisture transport)
  \item Food packaging (patterned channels to maintain freshness)
  \item Adaptive solar trackers (patterned curved panels)
  \item Solar cells (patterned like flower petals, circularly arranged)
\end{itemize} \\

What are other coatings that can freeze the dynamics of pollen paper? &
\begin{itemize}
  \item Paraffin wax
  \item Silica
  \item Carbon, graphene (to induce electrical properties)
  \item Magnetic particles (for remote shape control)
\end{itemize} \\
\bottomrule
\end{tabular}
\caption{Top ideas generated from prompts regarding the plant literature using the Idea Mining protocol. Ideas have been summarized for clarity and brevity.}
\label{tab:idea_mining_results}
\end{table}

\begin{table}[h]
\centering
\small 
\begin{tabular}{p{2.5cm} p{2.5cm} p{9cm}}
\toprule
\textbf{Prompt} & \textbf{Idea} & \textbf{Multi-Agent Summary} \\
\midrule
"What applications might \textit{Rhapis excelsa} leaves be used for?" 
& "Rhapis excelsa leaves can be used as a model for developing adaptive building materials" 
& Rhapis excelsa leaves can inspire adaptive building materials that enhance energy efficiency, structural integrity, and aesthetic appeal. The unique corrugated folding patterns can be replicated through strategies like computational modeling, material selection, modular design, robotic fabrication, maintenance planning, and adaptive systems. \newline

Key Benefits: \newline
- Energy Efficiency: Corrugated folding increases surface area-to-volume ratio, improving heat transfer and air circulation. \newline
- Structural Integrity: Hierarchical structures offer strength with minimal weight, ideal for roof trusses or wall panels. \newline
- Aesthetic Appeal: Unique geometries create dynamic facades and decorative elements. \\
\bottomrule
\end{tabular}
\caption{Example from the Idea Mining protocol showing a prompt, corresponding idea, and the multi-agent elaboration summary.}
\label{tab:idea_mining_agents}
\end{table}

Ideas generated by the system can be directly pursued in the laboratory. One illustrative example involves digital printing of pollen paper, a moisture-responsive material whose shape and actuation behavior can be programmed through patterned bilayers. In the original study, simple striped patterns were printed to induce coiling and folding ~\cite{Zhao2021DigitalMaterials}.

\begin{figure}[h]
    \centering
    \includegraphics[width=1\linewidth]{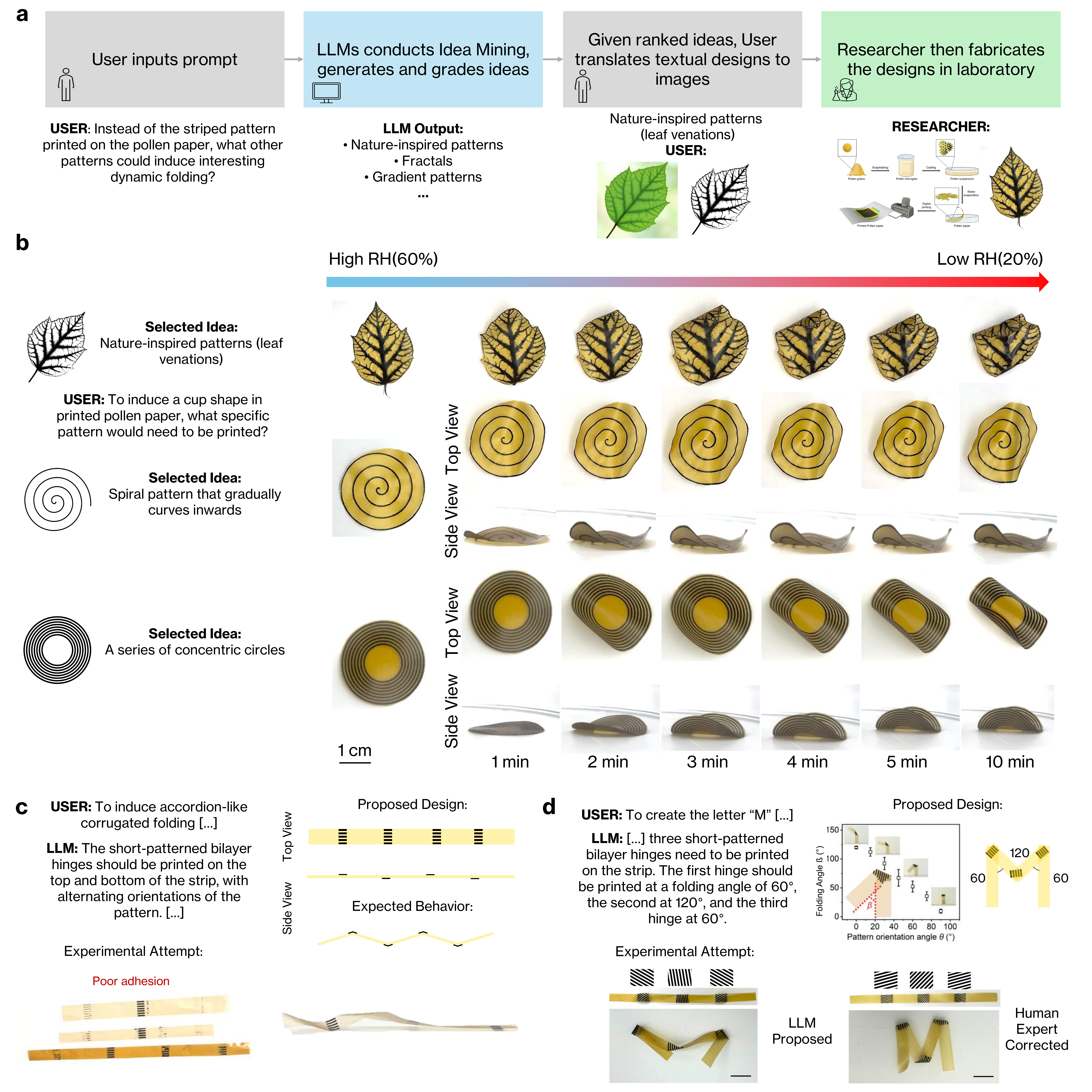}
    \caption{Fabricated \textit{de novo} materials designs in printed pollen paper a) Workflow from user input prompt to system generation and reasoning of ideas to the user generating image-based designs (the leaf image here is generated using an AI diffusion model lamm-mit/leaf-flux, a custom version of a FLUX.1 [dev] model fine-tuned using a image dataset of leaf microstructures) to experimental researchers fabricating the pollen paper and printing the designs. b) The humidity response of the resulting printed pollen paper designs when placed in high to low relative humidity (RH) conditions across time. c) Proposed design for corrugated folding actuation, followed by experimental attempt. Printing on both sides of pollen paper was not feasible due to poor toner adhesion, resulting in a limited humidity response. d) Proposed design for forming a letter "M" based on folding angle to pattern orientation guidelines (Image adapted from \cite{Zhao2021DigitalMaterials}. Licensed under
    CC BY-NC-ND 4.0). The initial experimental attempt using the LLM proposed design produced and incorrect, overly wide shape, which was later corrected by human experts. Scale bar is 1cm.}
    \label{fig:Fig_5}
\end{figure}

Extending upon this prior work, we used the system to generate \textit{de novo} pattern design ideas for printed pollen paper. As shown in Fig.~\ref{fig:Fig_5}a, the workflow begins with a user prompt asking for new pattern designs that could induce specific folding behaviors. The system runs the Idea Mining protocol to produce a ranked list of design concepts. The user then selects one or more ideas and creates corresponding visual representations of the patterns. These images are passed to experimental researchers, who fabricate the pollen paper and digitally print the designs onto it. 

Fig.~\ref{fig:Fig_5}b shows examples of the resulting paper materials and their humidity-responsive behavior when moved from high humidity (60\% relative humidity) to low humidity (20\% relative humidity) conditions, across time. One query led to the suggestion of a nature-inspired design, such as leaf venation patterns. The fabricated sample displayed folding behavior reminiscent of a dehydrated natural leaf. Another prompt asked for patterns that would induce a cup-shaped form. Two design concepts were selected, translated into images, and fabricated. The resulting samples showed measurable cupping, most apparent in side views where the edges lift upward.

However, the direct application of LLM proposed designs is not always seamless. As shown in Fig.~\ref{fig:Fig_5}c, the system is prompted to emulate corrugated folding in a strip of pollen paper, inspired by the folding behavior of \textit{Rhapis excelsa} leaves. The system proposes printing hinges on alternating sides of the paper. However, this overlooks a practical constraint known to human experts: the smooth side of the pollen paper does not retain toner well, making double-sided printing infeasible with the current standard of pollen paper. In the experimental attempt, hinges failed to adhere cleanly due to toner damage from high-temperature application on both sides. As a result, the strip exhibited limited folding actuation.

Fig.~\ref{fig:Fig_5}d shows another example in which the user prompts the system to form the letter "M" via hinge folding, following prior work that spelled out "NTU" using similar methods \cite{Zhao2021DigitalMaterials}. The system proposes a design based on folding angles, which appears theoretically sound. However, when translated into pattern orientation angles and fabricated, the resulting shape is incorrect as it is too wide at the center. Upon review, human experts adjusted the design by increasing the pattern orientation angle at the middle hinge to exaggerate the fold. This correction highlights a discrepancy between theoretical predictions and real-world behavior. Experimentalists note that pollen paper layers often require higher deformation angles than theory suggests, particularly due to material constraints and the cumulative effect of multiple hinges. They also emphasized that additional factors such as the paper’s weight and the non-uniform spatial distribution of pollen particles further influence actuation performance in practice. 

These two ‘unsuccessful’ examples offer valuable insights into the limitations of LLM-generated designs. In such cases, the LLM can also be reprompted to reflect on potential reasons for failure and propose corrective strategies, enabling a collaborative troubleshooting loop between the model and human expert. Still, while the outputs may appear directionally reasonable, they can overlook critical real-world nuances. Such cases underscore the importance of human tacit knowledge - practical design heuristics developed through hands-on experience, which are rarely captured in published literature. Since here, our LLM is fine-tuned on curated literature datasets that emphasize successful outcomes, they often miss the kinds of subtle constraints and failure modes that influence real-world performance. These observations underscore the need for the scientific community to more consistently document failed experiments, articulate design intuition and tacit knowledge, and provide detailed protocols, thereby enabling future AI systems to learn from a broader spectrum of outcomes.

Moreover, these examples demonstrate not only that LLM-generated ideas can be physically realized, but also reveal a key limitation: the need for human interpretation when translating text into fabrication-ready designs. Bridging this gap, potentially through multimodal AI systems, could further streamline and expand the capabilities of generative design pipelines in both 2D and 3D domains.

\subsection{Generating and executing AI-guided laboratory procedures}
\begin{figure}[h]
    \centering
    \includegraphics[width=1\linewidth]{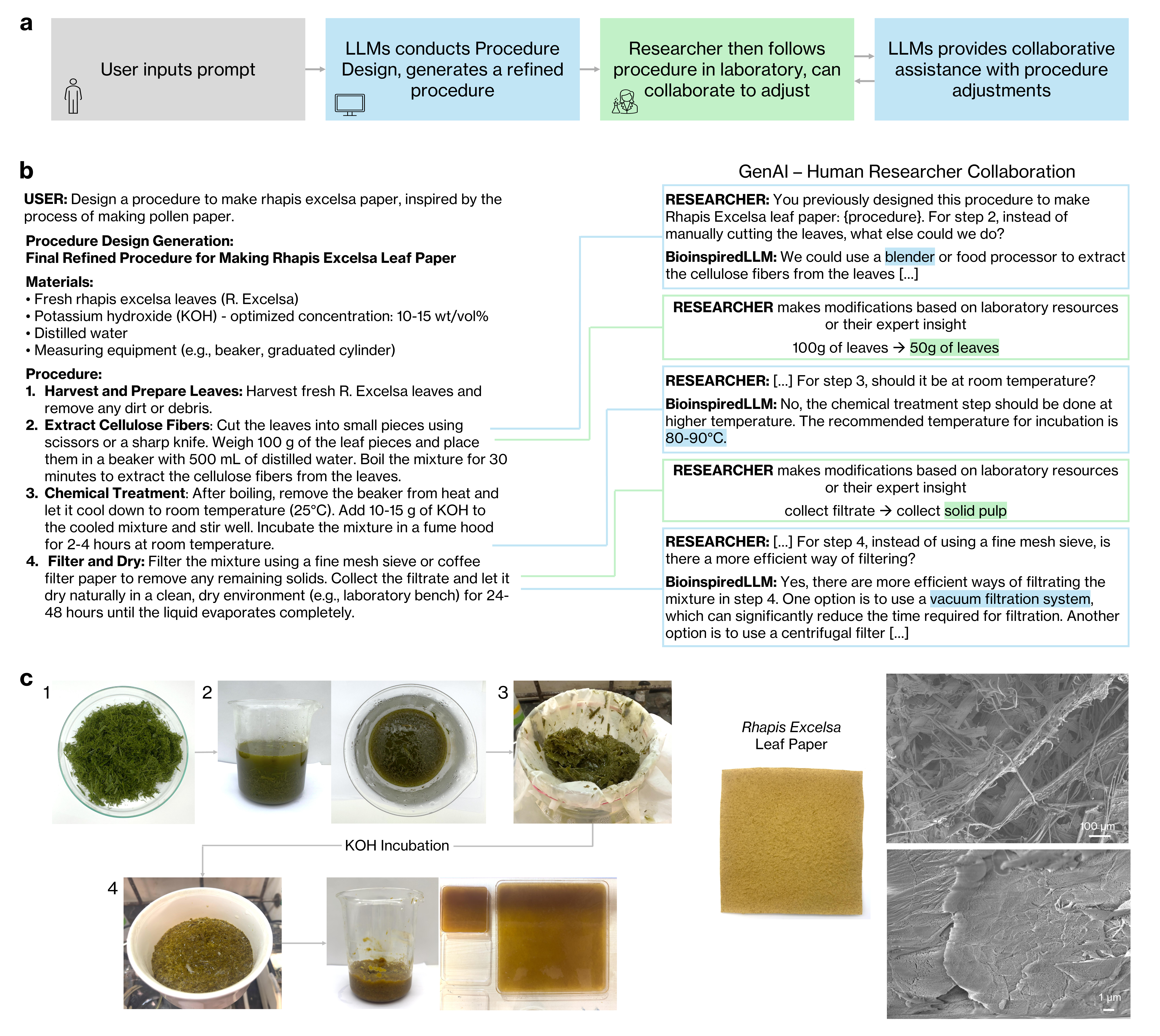}
    \caption{Human-AI execution of AI-generated procedure to make new materials. a) Workflow from user input prompt to system generation of procedure to then the collaboration of experimental researchers and the system to pursue the procedure physically. b) System generated procedure with annotations based on experimental researcher and AI discussion. c) Step by step image documentation of the fabrication process, ending with the final resulting paper and field emission-scanning electron microscopy images to document the resulting microstructure.}
    \label{fig:Fig_6}
\end{figure}

To generate actionable laboratory procedures for plant materials research, we implemented the Procedure Design protocol. One example involved the prompt: “Design a procedure to make \textit{Rhapis excelsa} leaf paper, inspired by the way pollen paper was made.” which has thus far not been applied to \textit{Rhapis excelsa}. The prompt was motivated by the potential for sustainable material design, drawing on prior work that transformed pollen grains into forms such as paper and cryogels. The resulting AI-generated procedure was experimentally validated in the laboratory. As shown in Fig.~\ref{fig:Fig_6}a, the full workflow begins with the user inputting a prompt into the system. The system then runs the Procedure Design protocol and outputs a detailed procedure. This output is reviewed and implemented by an experimental researcher, who may also interact with the LLM to make clarifying adjustments. Fig.\ref{fig:Fig_6}b presents the initial procedure generated by the system alongside annotated edits, which include insights from both further AI interactions and expert knowledge. Fig.\ref{fig:Fig_6}c documents the fabrication process step-by-step, with accompanying images. The final product - a sheet of \textit{Rhapis excelsa} paper - is shown with field emission-scanning electron microscopy (FE-SEM) images that reveal its resulting microstructure. While this specific proposed procedure is not scientifically novel and mirrors traditional methods, this experiment demonstrates the system’s ability to generate laboratory-ready protocols that are practical and implementable. More importantly, it highlights the collaborative potential between generative AI systems and experimental researchers, particularly in rapidly identifying viable starting points for real-world fabrication.

\begin{figure}[h]
    \centering
    \includegraphics[width=1\linewidth]{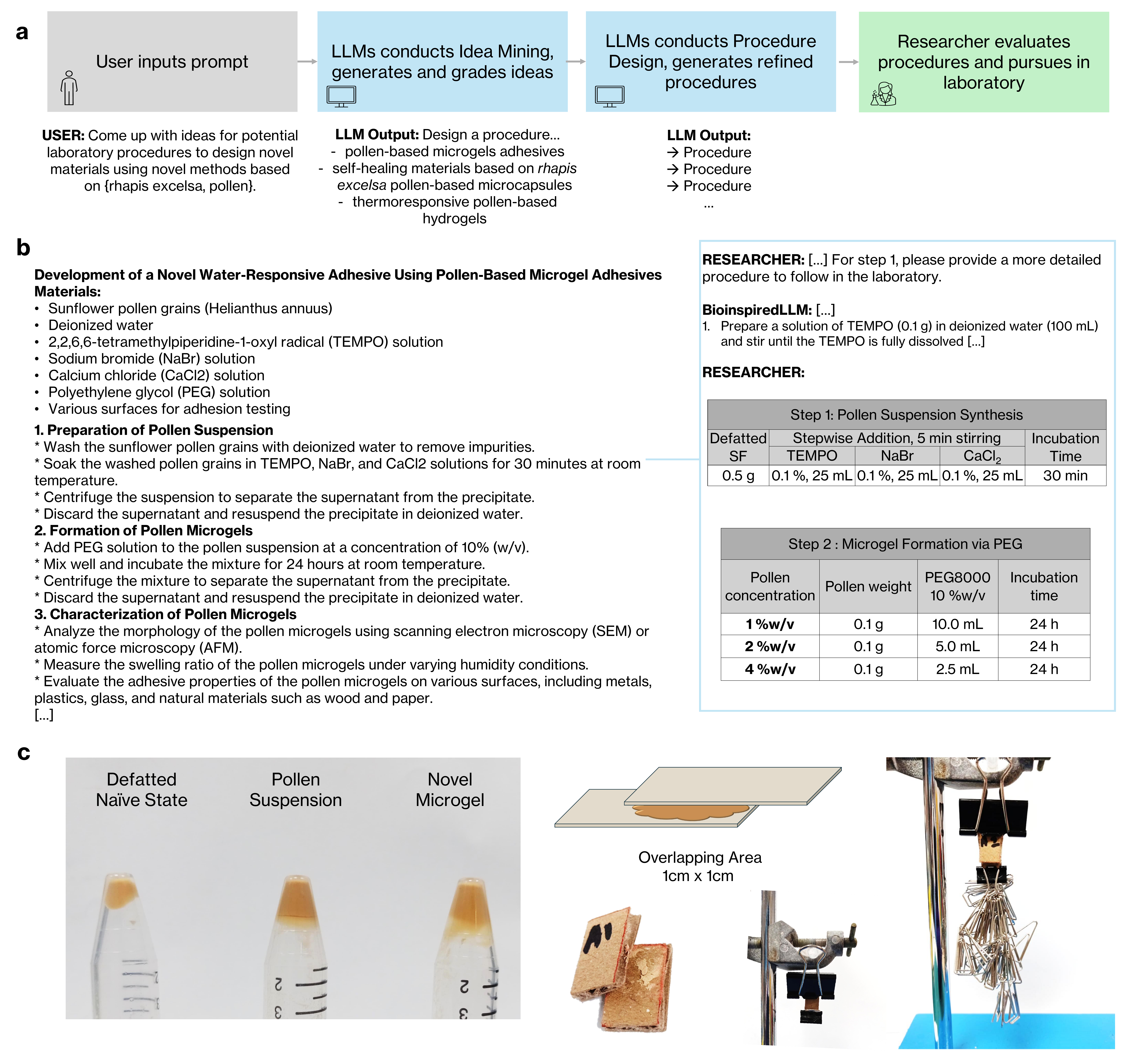}
    \caption{AI-guided scientific ideation from concepts to procedures. a) End-to-end workflow illustrating the transition from a user input prompt to AI-generated material design concepts, followed by AI-generated experimental procedures, and finally, collaboration between experimental researchers and the system to physically implement the selected procedure. b) Example AI-generated procedure for a novel water-responsive adhesive using pollen-based microgels, along with a researcher follow-up question used to refine specific parameters such as pollen concentration, measurement ratios, and processing details. c) Resulting pollen-based substances at each fabrication step. The final microgel exhibits a grainy texture and was preliminarily evaluated in a shear test, where it was applied between two cardboard substrates and its shear strength assessed via tensile loading.}
    \label{fig:Fig_7}
\end{figure}

While the previous example highlights the system’s ability to generate feasible, implementable procedures that support collaboration between AI and experimental researchers, we next sought to test the framework’s capacity for more open-ended innovation. Finally, we integrated all components of our framework to push the system toward generating the most novel and forward-looking designs for plant-inspired materials. As illustrated in Fig.~\ref{fig:Fig_7}a, we combined the Idea Mining and Procedure Design protocols to enable end-to-end generation—from conceptual ideation to experimental implementation.  In the Idea Mining phase, top-ranked concepts included pollen-based microgel adhesives, self-healing materials utilizing \textit{Rhapis excelsa} pollen microcapsules, and thermoresponsive pollen-based hydrogels. Each selected concept was then passed through the Procedure Design protocol to generate a corresponding fabrication procedure. Experimental researchers evaluated these procedures and ultimately selected the synthesis of water-responsive pollen-based microgel adhesives using a 2,2,6,6-tetramethylpiperidine-1-oxyl radical(TEMPO)-mediated oxidation process to pursue. Fig.~\ref{fig:Fig_7}b outlines the AI system's generated procedure along with clarifying adjustments made by human researches to refine specific parameters such as pollen concentration. The resulting pollen substances across each stage is shown in Fig.~\ref{fig:Fig_7}c, notably, the final microgel exhibits a grainy texture distinct from previously developed pollen-based microgels. These final formulations, prepared with varying pollen concentrations, were preliminarily evaluated in a shear test. In this test, the microgels were applied between two cardboard substrates, and shear strength was assessed via tensile loading. Among the tested formulations, the 2\%~w/v pollen concentration sample exhibited the highest adhesive shear strength. This example represents both a novel pollen-derived material and a previously unconsidered method for modifying pollen structures, one that had not been proposed by experimentalists, highlighting the remarkable potential of AI-assisted experimentation to surface unexpected and impactful directions for materials innovation.

\section{Conclusions}
In this work, we demonstrate that a structured generative AI system employing advanced inference strategies can accelerate progress in scientific and engineering research, moving beyond traditional single-shot LLM methods. By combining fine-tuned models (BioinspiredLLM), RAG, agentic systems, and the newly introduced Hierarchical Sampling approach, we present a vision for AI-assisted experimentation and hypothesis generation that is both robust and forward-looking. Beyond showcasing the methodological contributions of this framework, we apply it within the domain of plant-based materials to explore structure–property relationships and demonstrate how generative AI can support both ideation and real-world experimental realization. By rapidly generating, refining, and connecting ideas across domains, this system significantly reduces the time typically required for early-stage exploration and design iteration.

Importantly, we validate the framework not only through human and automated evaluations but, crucially, through experimental implementation in the laboratory. One compelling outcome is the generative design of a novel pollen-based adhesive, which originated from AI-guided ideation and was subsequently validated through laboratory fabrication. This end-to-end result demonstrates the system’s potential to contribute to practical bioinspired applications. We show that AI-generated hypotheses can translate into observable material behaviors, highlighting the feasibility and impact of integrating LLM-based systems into materials research pipelines.

However, as demonstrated, several active challenges remain for LLM-based systems. While the system is capable of generating directionally reasonable designs and can often predict correct experimental outcomes, it is not infallible and may overlook critical real-world constraints, particularly those arising from unwritten knowledge, material-specific limitations, or undocumented design heuristics. These limitations reflect broader gaps in current training paradigms, which largely rely on curated literature that favors successful outcomes and omits failure cases. Addressing these challenges will require greater access to negative results, intuitive design rationales, and detailed experimental protocols. Such practices would not only support reproducibility, but also provide a richer foundation for training AI systems capable of more robust and context-aware design.

This study focuses specifically on humidity-responsive plant materials, particularly pollen-based systems and \textit{Rhapis excelsa} leaf mechanics. Future work may expand this dataset using automated text and data mining, unlocking deeper insights across broader biological and engineered material systems. More broadly, this work demonstrates not only a rigorous investigation into plant mechanics, but also a generalizable, modular framework for AI-assisted materials discovery. While generative AI holds tremendous potential, realizing its full value requires interpretability and accessibility. Previous efforts have explored complex multi-agent architectures involving three or more models; in contrast, we present a streamlined, computationally accessible approach using just two quantized agents, making it practical for researchers without extensive AI expertise. All code, prompts, and notebooks are provided to support reproducibility and future development. While the overall system may appear complex, we offer modular, well-documented code designed to be flexible, adaptable, and easily extended to new use cases. Our developed protocols further support this by producing structured outputs that help researchers derive actionable and impactful insights.

Finally, this work highlights the potential of a collaborative partnership between generative AI and human expertise. The system can accelerate ideation, synthesize prior knowledge, and surface promising hypotheses, including ideas not yet explored in the literature. While some of these ideas may not have been proposed by researchers before, this does not necessarily indicate that they are beyond human reach. Rather, they may emerge more efficiently through the model's ability to systematically recombine and explore a broad range of concepts at scale. At the same time, human insight remains essential in interpreting outputs, refining designs, correcting model errors - such as those involving precise shape generation - and navigating real-world constraints. Ongoing advances in multimodal and spatially aware models may help close some of these gaps. As these tools continue to evolve, their ability to contribute more autonomously to design and experimentation will likely expand. Together, these complementary strengths point toward a future in which researchers and AI work side by side by side to unlock new frontiers in plant and materials science.

\section{Materials and methods}
\label{sec:Materials_Methods}
 
We provide details on the materials and methods used to conduct this study. 

\subsection{LLMs and RAG}

BioinspiredLLM, first developed in \cite{Luu2023BioinspiredLLM:Materials}, here is based on the most recent Llama-3.1-8b-instruct architecture, specifically the model weights can be found on HuggingFace repository \texttt{lamm-mit/Llama3.1-8b-Instruct-CPT-SFT-DPO-09022024}. This specific version of BioinspiredLLM, was developed in \cite{Lu2024Fine-tuningCapabilities_updated} using continued pre-training, supervised fine-tuning, and direct preference optimization (DPO). The second model used is the base model,  \texttt{meta-llama/Llama-3.1-8b-instruct}. Both models were quantized via GPT-Generated Unified Format
(GGUF) files to Q8\_0 bit for single LLM inference as well as Q4\_K\_M for multi-agent tasks to be employed on modest hardware. Models were loaded using a Python binding of  LlamaCPP integrated with LlamaIndex. LlamaIndex was also used for query engine for RAG, using the HuggingFaceEmbedding  \texttt{BAAI/bge-small-en-v1.5 model}. The simple directory reader was employed to load PDF files of the plant literature. 

To optimize BioinspiredLLM equipped with RAG, a sensitivity analysis was performed to adjust key RAG parameters, including node length and top-k values. The complete results of this analysis are documented in detail in Supplementary Information Section 1.1. 

\subsection{Agentic Systems}
Multi-agent conversations were facilitated using a custom script enabling multi-turn conversations between BioinspiredLLM and Llama-3.1-8b-instruct. In each interaction, the system prompts of each model are tailored to roles based on the protocol. The user initiates the first turn of conversation as well as modulates the number of turns, in this case only 2-3 in the experiments presented here. After the exchange, Llama-3.1-8b-instruct summarizes the conversation into a single distilled output, or, in the case of procedure design, a final procedure. More details regarding the exact system prompts are documented in detail in Supplementary Information.

\subsection{Hierarchical Sampling and Prompts}
For brevity, all prompts used are documented in detail in the Supplementary Information. Prompts were crafted based on effective performance and tailored to expert interests in specific applications and topics.

Hierarchical Sampling was implemented using a custom recursive script that enables repeated LLM sampling to generate hundreds (or any user-specified amount) of unique ideas. Diversity filtering, ranking, and refinement were handled within the same pipeline. All scripts, including those used for prompts and protocol execution, are available for reproducibility and future adaptation.

\subsection{Evaluation Techniques}
As described in the study, a comparative analysis of our system and an off-the-shelf LLM (Llama-3.1-8b-instruct) was conducted using a combination of human-expert evaluation as well as automated assessment through an uninvolved foundational LLM, in this case (GPT-4o) \cite{achiam2023gpt}. Automated assessment via LLM is a relatively established procedure as assessed in previous multi-task natural language processing settings \cite{liang2022holistic, chiang2023can, zhang2024llmeval, chern2024can}. For the Idea Mining experiment, the following metrics were used as the rubric:
\begin{itemize}
    \item Creativity: The novelty and innovation of the ideas.
    \item Uniqueness: How different and varied the ideas are within its own list. Reduce ratings for redundancy.
    \item Specificity: The extent to which the ideas capture niche or specialized concepts. Reduce rating for focusing on too broad or basic concepts.
\end{itemize}

Additionally, the novelty score was calculated by GPT-4o equipped with Semantic Scholar access, similarly to \cite{Ghafarollahi2024SciAgents:Reasoning}. The task is first designed to examine an idea and decide on 3-5 keywords relevant to the idea. Those keywords are called in Semantic Scholar with the top 10 relevant results (if any) to be retrieved as the model decides on the novelty score.

For the Procedure Design experiment, the following metrics were used as the rubric:
\begin{itemize}
    \item Novelty: Likewise, aided using Semantic Scholar literature search to capture the procedure methodology
    \item Specificity: Measured the degree of detailed technical description in generated procedures.
    \item Tractability: Measured how realistically the procedures could be implemented in the lab, including completeness and feasibility.
\end{itemize}

Complete prompts used for the LLM evaluation are documented in detail in Supplementary Information.

In blinded human reviews, evaluators were given similar rubrics. These metrics were used specifically to determine which procedures were most promising to pursue in the laboratory, based on process novelty (the originality of the approach), specificity (the level of detail and clarity), and tractability (the feasibility and completeness of the proposed steps for real-world implementation).

\subsection{Laboratory Experiments}
All laboratory experiments were conducted at Nanyang Technological University in Singapore. For experimental fabrication, standard pollen processing procedures were followed as previously documented \cite{Zhao2020ActuationPollen, Zhao2021DigitalMaterials,Deng2024Plant-BasedControl}, except in cases where AI-generated suggestions introduced modifications or new protocols.

\section*{Acknowledgments}
This work was supported in part by Google, the MIT Generative AI Initiative, USDA (grant number 2021-69012-35978), with additional support from NIH. This material is based upon work supported by the National Science Foundation Graduate Research Fellowship under Grant number 2141064. This research was supported in part by National Science Foundation under grant number DMR-2004556. This work was supported by Nanyang Technological University, Singapore and funded by Ministry of Education (MOE, Singapore) Academic Research Fund Tier 3 2022 under Award Number MOE-MOET32022-0002. The work was also funded by Global Green Growth Institute (REQ0412626, Funding Number: 023138-00001) and Ministry of Agriculture, Food and Rural Affairs KIPETFAF (RS-2024-00403067).

\section*{Author contributions}
R.K.L. conceived the sampling strategies, developed inference algorithms and code, and conducted all digital assessments and analyses. R.K.L. also wrote the initial draft of the manuscript. M.J.B., S.S., M.D., and N.-J.C. supervised the project and contributed to its conceptual development. J.D., M.S.I., and N.-J.C. analyzed and evaluated the output procedures, and provided consultation to the research. J.D. and M.S.I. designed and performed the benchtop experiments. All authors contributed to editing and finalizing the manuscript.

\section*{Competing interests}
The authors declare no conflicts of interest.

\section*{Code and model weight availability}

All code, protocols, and notebooks developed in this study are available at:
\url{https://github.com/lamm-mit/LLMsxPlants}. Model weights can be accessed at:
\url{https://huggingface.co/lamm-mit}.

\bibliographystyle{naturemag}
\bibliography{references-mendeley, references}  

\clearpage


\section*{Supplementary material}

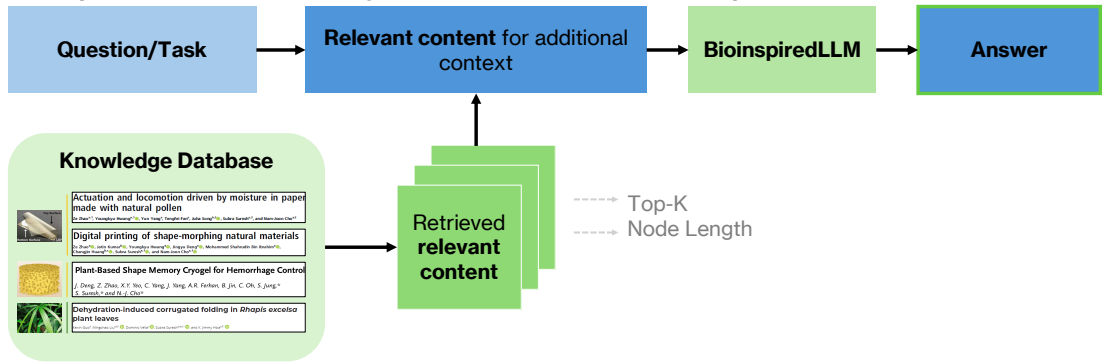

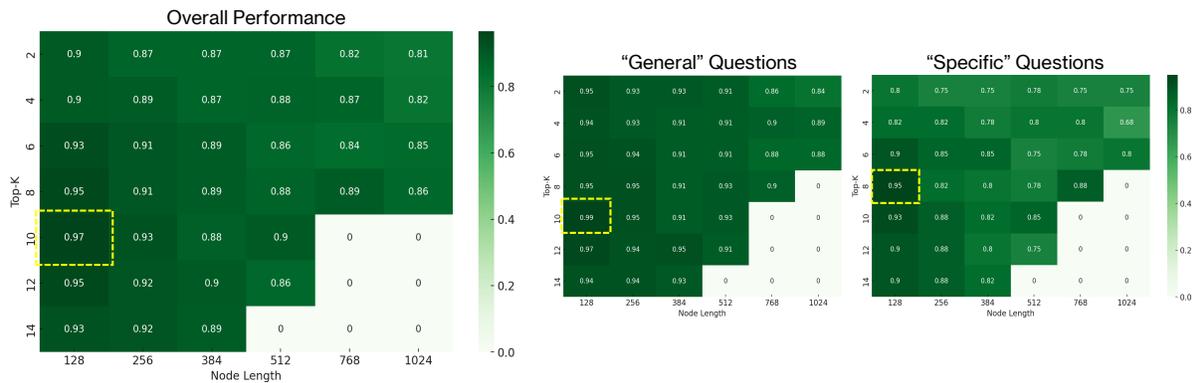

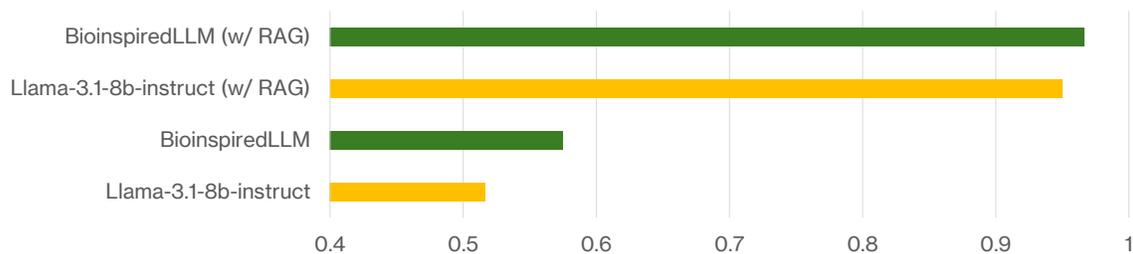

Figure 1: a) Overview of BioinspiredLLM w/RAG Set-Up, highlighting the knowledge database consisting of the plant literature. b) Optimization of RAG parameters, top-k and node length. From left to right, for overall performance, then separated by "General" questions, and "Specific" questions. Yellow dotted boxes highlight the top performing parameter scenarios in each case. c) Knowledge benchmark performance across model candidates including BioinspiredLLM w/RAG, Llama-3.1-8b-instruct w/RAG, BioinspiredLLM, and Llama-3.1-8b-instruct.

The benchmark was developed using methods similar to those outlined in previous work [1]. It comprises technical multiple-choice questions ranging from broad, general trends to highly specific queries addressing key concepts or numerical data analysis. The benchmarks were used to establish baseline performance of the models and to optimize RAG parameters. Using developed benchmarks specific to the plant literature, the results are summarized in Fig. 1b. These benchmarks included both "General" and "Specific" questions. General questions captured broader trends and themes in the plant literature, while specific questions focused on distinct details or numerical results from the papers.

The optimized parameters consisted of a top-k value of 10, meaning that the system retrieves the 10 most relevant text chunks from the knowledge database before generating a response. A node length of 128 was used to define the size of each text segment stored in the retrieval index, ensuring that retrieved information contains sufficient context

while remaining computationally efficient. Additionally, a token overlap of 50 was applied, meaning that 50 tokens from the end of one segment were repeated in the next segment to maintain continuity and avoid loss of contextual meaning during retrieval. These parameters struck a balance between providing sufficient context and avoiding excessive information that could dilute relevance. Notable differences emerged between the performance on General and Specific questions. For specific questions, the optimal top-k value was slightly lower, and performance showed a greater decline with increasing node lengths. This suggests that an excessive amount of RAG-generated context can negatively impact the model's ability to identify specific details, highlighting the importance of fine-tuning RAG parameters based on the nature of the task.

After optimizing the RAG parameters, the models were tested, and their performance is illustrated in Fig. 1c, comparing scenarios with and without RAG. BioinspiredLLM consistently outperforms its base model, Llama-3.1-8b-instruct, in both conditions. Without RAG, the performance is subpar but still generally exceeds that of Llama-3.1-8b-instruct, likely due to BioinspiredLLM's enhanced familiarity with biological materials. With RAG, BioinspiredLLM demonstrates stronger results, although the performance of Llama-3.1-8b-instruct with RAG is also notably improved, indicating the substantial impact of RAG on foundational model capabilities. These findings underscore the knowledge retention and retrieval potential of foundational models, particularly when equipped with RAG. Given these results, Llama-3.1-8b-Instruct was integrated into the proposed multi-agent protocols for knowledge retrieval tasks, while BioinspiredLLM was used for both knowledge retrieval and creative divergent generation tasks.

### 1.2 Agentic Systems

The following system prompts for the agentic systems are as follows for the Idea Mining and Procedure Design protocols, respectively:

---
**System Prompts: Idea Mining - Multi-Agents**

**BioinspiredLLM - SYSTEM PROMPT:**
```
You are a BioinspiredLLM, a creative engineer with knowledge in biological and bio-inspired materials. You are taking
part in a discussion and have access to potentially relevant information from previous papers on Rhapis excelsa
leaves and pollen-based materials.  However, do not cite studies or references. Focus on providing concise, accurate,
and thoughtful contributions that stimulate new trains of thought.
```

**Llama-3.1-8b-instruct - SYSTEM PROMPT:**
```
You are a Engineer, taking part in a discussion where your role is to ask concise, challenging questions to evaluate
effectiveness and feasibility. Do not ask for specific studies, sources, or references. Instead, prioritize
reasoning, logical connections, and thoughtful consideration of potential implications or improvements.You have
access to potentially relevant information from previous papers on Rhapis excelsa leaves and pollen-based materials.
```
---

**System Prompts: Procedure Design - Multi-Agents**

**BioinspiredLLM - SYSTEM PROMPT:**
```
You are a BioinspiredLLM, a creative engineer with knowledge in biological and bio-inspired materials. You are taking
part in a discussion and have access to potentially relevant information from previous papers on Rhapis excelsa
leaves and pollen-based materials.  However, do not cite studies or references. Focus on providing concise, accurate,
and thoughtful contributions that stimulate new trains of thought.
```

**Llama-3.1-8b-instruct - SYSTEM PROMPT:**
```
You are a Scientist, taking part in a discussion where your role is to collaborate with BioinspiredLLM to develop a
scientific procedure. Prioritize scientific effectivenes and logical reasoning when providing potential implications
or improvements.You have access to potentially relevant information from previous papers on Rhapis excelsa leaves and
pollen-based materials.
```
---

### 1.3 Prompts

#### 1.3.1 Mechanistic Insights

For the correlation of mechanical properties to plant structures to engineered structures task, the following prompt algorithm was developed to rapidly generate and reason across many designs and mechanical properties of interest. For each prompt that asks the model to make a list, the generated output is extracted line by line, with each line sampled as the next query.

---

**Prompt Algorithm: From Engineering Properties to Plant Inspired Designs**

**User:** List the 10 most important structures and architectures found in pollen grains or pollen-based materials that influence its {property}. Provide the list in bullet point format, without any explanations or additional context. List only 10 or less.

**BioinspiedLLM:** ...

**User:** Describe concisely how {structure} may contribute to {property} in pollen grains or pollen-based materials. Focus on its primary mechanism or role, and critically assess its significance. No citations."

**BioinspiedLLM:** ...

**User:** The following is a list of universal structures in pollen grains or pollen-based materials and how they contribute to {property}: {structure} + {rationale}
Based on their significance, identify the top {top_n} structures that most strongly contribute to {property}. Ensure that you select physical structures, not properties.
Provide the names of the top {top_n} structures in a bullet point list, and include a one-sentence rationale for why each structure was selected.

**BioinspiedLLM:** ...

**User:** Here is a structure of pollen grains or pollen-based materials, its mechanism, and the property of interest:
    Structure: {structure}
    Mechanism: {rationale}
    Property: {property}

Based on this information, propose an engineered structure that mimics the structure and mechanism, optimized for {property}. Focus on the geometry or shape of the engineered structure. Be concise and specific.

**BioinspiedLLM:** ...

---

Likewise, a similar algorithm was designed for the structure-property relationship extraction task. Again, for each prompt that asks the model to make a list, the generated output is extracted line by line, with each line item sampled individually as part of the next query.

---

**Prompt Algorithm: Identification of Structure-Property Relationships**

**User:** List the top {top_n} intrinsic mechanical properties of pollen. Provide the list in bullet point format, without any explanations or additional context.

**BioinspiedLLM:** ...

**User:** Describe concisely how the mechanical property, {property}, is influenced in pollen. Focus on its primary mechanism or role, and critically assess its significance. No citations.

**BioinspiedLLM:** ...

**User:** The {mech_property} is a property of pollen. Explanation: {explanation}.
List the intrinsic structures or architectures found in pollen that affect {mech_property}.Provide the list in bullet point format with names only, with no explanations or additional context. Limit to 10 structures.

**BioinspiedLLM:** ...

**User:** The following is a list of structures found in pollen that may contribute to {property_name}: {structure_details}
Identify the top {top_n} structures or architectures that most strongly contribute to {property_name}. Provide the names of the top {top_n} structures in a plain bullet point list, followed by a brief one sentence rationale

**BioinspiedLLM:** ...

---

The results are extracted and stored as JSON outputs. For the second task, the bipartite graph is constructed in Python using Matplotlib.

### 1.3.2 Idea Mining

The following are the detailed prompts for the idea mining protocol. For each phase, a .csv file is outputted with the results for easy readability.

> **Prompt: Idea Mining Divergent Generation**
>
> **User:** `{prompt} Be creative. Concisely brainstorm {N} different ideas into a bullet point list.`
>
> **BioinspiredLLM:** ...
> **BioinspiredLLM:** ...
> **BioinspiredLLM:** ...
>
> `[sampled, filtering out duplicate using cosine similarity, until final list of ideas > N]`

Two methods of convergent evaluation were conducted. The ratings method is where the LLM first gives ratings to all of the ideas based on novelty and effectiveness, passed two random ideas at a time. The pairwise elimination method is where the LLM does pairwise elimination of ideas, given randomized sets of two ideas, for a series of rounds, leading to a ranked list of ideas. The following are the prompts used, omitting parts that dictate the LLM to output in JSON format for brevity.

> **Prompt: Idea Mining Convergent Evaluation - Ratings**
>
> **User:** `A brainstormed list of ideas was created to answer this question: {prompt}. You must critically rate each idea out of 10 for categories: novelty and effectiveness.`
>
> `Idea 1: {idea1}`
> `Idea 2: {idea2}`
>
> **Llama-3.1-8b-instruct:** ...

> **Prompt: Idea Mining Convergent Evaluation - Pairwise Elimination**
>
> **User:** `To answer this question: {prompt} Which idea is better based on novelty and effectiveness?`
> `Idea 1: {idea1}`
> `Idea 2: {idea2}`
>
> **Llama-3.1-8b-instruct:** ...
>
> `[continued until user-specified number of rounds is reached]`

Upon generation of ideas and evaluations, the user may select ideas of interest to pass to a multi-agent conversation, in which it follows the following multi-agent system prompts and prompts. The final output is both the whole multi-agent conversation as well as the LLM-generated summary/final answer of the conversation.

> **Prompt: Idea Mining - Multi-Agent**
>
> **User:** `Explore this new idea further: {idea}`
>
> **BioinspiredLLM:** ...
>
> **Llama-3.1-8b-instruct:** ...
>
> **BioinspiredLLM:** ...
>
> **Llama-3.1-8b-instruct:** ...
>
> `[continued until number of user-specified turns are completed]`
>
> **User:** `Carefully read this conversation. Develop a final answer to the original question using the conversation as additional context.`
>
> **Llama-3.1-8b-instruct:** ...

### 1.3.3 Procedure Design

The generation of Questions and Answers (Q-As) for the procedure design protocol is as follows:

---

**Prompt: Procedure Design - Q-As**

**User:** `In order to '{prompt}', create a concise list of very basic and fundamental questions that explore the essential properties, definitions, and background relevant to this topic.`

**BioinspiredLLM:** ...

`[questions are extracted as a list and passed individually into the next query]`

**User:** `{question}. Answer concisely and accurately. If you don't know the answer or there isn't enough context from the provided information, state that this area needs further exploration. Do not use citations.`

**BioinspiredLLM:** ...

---

For the multi-agent development of the procedure, the following multi-agent system prompts and prompts were used. The final output is both the whole multi-agent conversation as well as the LLM-generated final procedure based on the conversation.

---

**Prompt: Procedure Design - Multi-Agent**

**User:** `Design a procedure to {prompt}. Here is some additional information that can support this task {Q-As}.`

**BioinspiredLLM:** ...

**Llama-3.1-8b-instruct:** ...

**BioinspiredLLM:** ...

**Llama-3.1-8b-instruct:** ...

`[continued until number of user-specified turns are completed]`

**User:** `Carefully read this conversation. Outline the final, refined procedure using the conversation as additional context.`

**Llama-3.1-8b-instruct:** ...

---

### 1.3.4 LLM Evaluation

As described in the main text, a third party LLM was used to conduct additional evaluation.

---

**Prompt: LLM Creativity Evaluation**

**User:** `For the prompt {prompt}, two lists of around 100 ideas each were created. Your task is to be very critical. Compare both of the lists and rate the quality of each list's ideas based on three categories: 'creativity,' 'uniqueness,' and 'specificity'. Each category should be rated on a scale from 1-100, where 100 is the highest rating.`
`**Definitions:**`
`    - **Creativity**: The novelty and innovation of the ideas.`
`    - **Uniqueness**: How different and varied the ideas are within its own list. Reduce ratings for redundancy.`
`    - **Specificity**: The extent to which the ideas capture niche or specialized concepts. Reduce rating for focusing on too broad or basic concepts.`

`List 1: {list1}`

`List 2: {list2}`

**LLM:** ...

---

Additionally, the novelty score was calculated by GPT-4o equipped with Semantic Scholar access, similarly to [2]. The task is first designed to examine an idea and decide on 3-5 keywords relevant to the idea. Those keywords are called in Semantic Scholar with the top 10 relevant results (if any) to be retrieved as the model decides on the novelty score. The following prompt algorithm is shown below:

> **Prompt: LLM Semantic Scholar Novelty Evaluation**
>
> **User:** The following research idea was generated in response to this prompt: '{prompt}'.
> Research Idea: "{idea}"
> Generate three keywords that best summarize the core concepts of this idea. These keywords will be used in a literature search engine to find relevant papers, so make sure to include specific terms (like names, technical terms, or key concepts) that are essential for retrieving precise results.
>
> **LLM:** [Keyword Generation followed by Semantic Scholar API search for top 10 entries, including paper title and abstract]
>
> ---
>
> **User:** You are a critical scientist assessing the novelty of research ideas. In response to this prompt: "{prompt}", the following idea was generated: "{idea}".
>
> Your primary task is to evaluate this idea for novelty on a scale of 1-100, ensuring it does not significantly overlap with existing literature or cover areas that are already well-explored.
>
> To aid your assessment, here are the top 10 results from literature search using the keywords:'{keywords}':
>
> {semanticscholaresults}
>
> Based on your analysis, rate the novelty of this idea strictly and respond only with a number from 1 to 100.
>
> **LLM:** ...


 
\end{document}